\documentclass[letterpaper, 10 pt, conference]{ieeeconf}  

\IEEEoverridecommandlockouts                              

\usepackage{graphicx} 
\usepackage{times} 
\usepackage{amsmath} 
\usepackage{amssymb}  
\usepackage{url}
\usepackage{array}
\usepackage{cite}
\usepackage[hidelinks]{hyperref}
\usepackage{tablefootnote}
\usepackage{booktabs}
\title{\LARGE \bf
Motion Control in Multi-Rotor Aerial Robots Using Deep Reinforcement Learning
}
\author{Gaurav Shetty$^{1,2}$, Mahya Ramezani$^{1}$, Hamed Habibi$^{1}$, Holger Voos$^{1}$, and Jose Luis Sanchez-Lopez$^{1}$
\thanks{$^{1}$Authors are with the Automation and Robotics Research Group, Interdisciplinary Centre for Security, Reliability, and Trust (SnT), University of Luxembourg, Luxembourg. Holger Voos is also associated with the Faculty of Science, Technology and Medicine, University of Luxembourg, Luxembourg. \texttt{\{gaurav.shetty, mahya.ramezani\}@ext.uni.lu},
\texttt{\{hamed.habibi, holger.voos, joseluis.sanchezlopez\}@uni.lu}}%
\thanks{$^{2}$Gaurav Shetty is also with the Bonn-Rhein-Sieg University of Applied Sciences, Germany.}
}

\begin{document}

\maketitle
\thispagestyle{empty}
\pagestyle{empty}

\begin{abstract}

This paper investigates the application of Deep Reinforcement Learning (DRL) to address motion control challenges in drones for additive manufacturing (AM). Drone-based additive manufacturing offers a flexible and autonomous solution for material deposition in large-scale or hazardous environments. However, achieving robust real-time control of a multi-rotor aerial robot under varying payloads and potential disturbances remains challenging. Traditional controllers like PID often require frequent parameter re-tuning, limiting their applicability in dynamic scenarios. We propose a DRL framework that learns adaptable control policies for multi-rotor drones performing waypoint navigation in AM tasks. We compare Deep Deterministic Policy Gradient (DDPG) and Twin Delayed Deep Deterministic Policy Gradient (TD3) within a curriculum learning scheme designed to handle increasing complexity. Our experiments show TD3 consistently balances training stability, accuracy and success, particularly when mass variability is introduced. These findings provide a scalable path toward robust, autonomous drone control in additive manufacturing.

\end{abstract}

\section{Introduction}

Unmanned Aerial Vehicles (UAVs) constitute a technology within autonomous systems, with applications in precision agriculture, infrastructure inspection, and logistics \cite{c1}. Recently, the integration of UAVs with Additive Manufacturing (AM) capabilities has been recognized as a promising development \cite{c2}. UAV-based AM combines the aerial mobility of drones with 3D printing technologies, enabling material deposition in complex or otherwise inaccessible environments \cite{c3}. This integration introduces new possibilities for construction and repair, particularly for large-scale structures that are difficult to access using traditional methods.

However, the deployment of UAVs for AM tasks presents control challenges \cite{c4}. As drones deposit materials, their mass distribution changes continuously, altering both the center of gravity and inertia. In addition, external disturbances further complicate control, making it difficult to maintain stability and accuracy in dynamic three-dimensional environments \cite{c5}. These challenges highlight the need for robust and adaptive control systems \cite{c27}.

Traditional control systems \cite{c6,c7_new}, have been widely utilized in UAV navigation. These methods rely on fixed parameters and predefined models, performing effectively in predictable environments \cite{c7,c9_new}. However, within the context of UAV-based AM, the continuous changes in mass distribution and the presence of external forces require frequent recalibration of these controllers. This necessity limits their suitability for real-time adaptive applications, where dynamic and unpredictable conditions are prevalent \cite{c8}.

Reinforcement Learning (RL) offers an alternative approach to UAV control by enabling systems to learn control policies through interaction with their environment, without depending on explicit system models. This model-free method allows RL algorithms to adapt dynamically to both internal variations and external disturbances  \cite{c9}. However, standard RL methods encounter challenges in high-dimensional environments like AM, where state and action spaces are extensive and the environment is highly unpredictable \cite{c10, c26}.

Deep Reinforcement Learning (DRL) builds on traditional reinforcement learning by using neural networks to approximate value functions and policies. This approach enables effective learning in high-dimensional state and action spaces. DRL proves effective in addressing the complex challenges associated with UAV-based AM \cite{c11}. 

A key consideration in applying DRL to UAV-based AM is the choice between deterministic and stochastic policy methods. Deterministic approaches, such as Deep Deterministic Policy Gradient (DDPG) \cite{c12}, produce specific actions for a given state. This ensures higher precision and consistency \cite{c13}, which are essential for accurate material deposition in AM. Stochastic methods sample actions from probability distributions, introducing variability that can reduce precision. Deterministic methods are better suited for UAV-based AM as they provide predictable and repeatable control. They also reduce computational overhead and work well in stable, low-noise environments \cite{c14}. However, DDPG faces challenges such as instability and overestimation bias \cite{c9}. Twin Delayed Deep Deterministic Policy Gradient (TD3) \cite{c15} addresses these issues. It uses twin critics to reduce overestimation and delayed policy updates to improve stability. These improvements make TD3 an effective choice for UAV-based AM tasks that require precise and reliable control under varying conditions \cite{c16}. 

Despite advancements, applying DRL to UAV-based AM faces several practical issues. These include improving policy stability during training, adapting to dynamic task complexity, and ensuring reliable performance under changing operational conditions \cite{c17}. Curriculum learning helps address these issues by using a gradual training strategy. It starts with simple navigation tasks and incrementally introduces dynamic waypoints, variable payloads, and external disturbances. This staged approach stabilizes training, improves robustness, and enhances real-world applicability \cite{c18}. 

Waypoint-based navigation simplifies the control problem by dividing it into discrete target points. It provides clear evaluation metrics, measuring accuracy through the average positional error and precision via the standard deviation of these errors. This approach enables systematic testing of control algorithms, facilitates benchmarking in dynamic environments, and ensures precise assessments of UAV control strategies \cite{c19}. 

This work focuses on controlling a UAV in 3D for AM using DDPG and TD3. Curriculum learning is integrated to improve training stability and performance under increasing task complexity. The study compares DDPG and TD3 to identify the most effective control strategy for UAV-based AM.

Key contributions include:
\begin{itemize}

\item The UAV-AM control problem is modeled as a Markov Decision Process (MDP) \cite{c20} with well-defined states, actions, rewards, and constraints. The state includes accelerations in the x, y, and z directions, to allow the agent to infer mass variations and adapt thrust commands.
\item The performance of DDPG and TD3 is evaluated using metrics such as average cumulative reward, positional error and precision. 
\item Curriculum learning is introduced to enable the agent to handle progressively complex tasks effectively. 
\item Waypoint-based navigation is used to simplify the control problem and facilitate evaluation and training. 

\end{itemize}

This study provides a framework for improving UAV control in AM, supporting advancements in autonomous construction and repair technologies.

\section{Problem Formulation And Preliminaries}

This study focuses on developing a precise and stable UAV control system for AM in a structured environment. The UAV operates in a three-dimensional space governed by nonlinear dynamics, incorporating thrust forces, gravitational effects, and external disturbances. The mathematical formulation of the UAV’s motion dynamics follows standard multirotor modeling, as detailed in \cite{mellinger2012trajectory}. 

The control objective is to enable accurate waypoint navigation while adapting to system variations. To simplify the problem, waypoints are predefined in a fixed sequence, eliminating the need for path planning and allowing a focus on control adaptation. Material deposition is modeled by introducing a reaction force in the negative z-direction to simulate the UAV counteracting deposition forces. The force exerted on the UAV due to material deposition is given by  

\[
F = \dot{m} v_{\text{exit}},
\]

where \( \dot{m} \) is the mass flow rate of the extruded material and \( v_{\text{exit}} \) is the velocity of material leaving the nozzle. Based on the material properties reported in \cite{zhang2022aerial}, with a density of \( 1700 \) kg/m³, a nozzle diameter of \( 8 \) mm, and a flow velocity of \( 0.5 \) m/s, the calculated force is approximately  

\[
F = 2.135 \times 10^{-2}~\text{N}
\]

The corresponding acceleration imparted to the UAV is computed as  

\[
a = \frac{F}{m},
\]

where \( m \) is the UAV mass. This additional acceleration is incorporated into the UAV’s motion model in the simulation to effectively account for the impact of material deposition. Additionally, random noise is introduced into sensor measurements to ensure robustness against real-world uncertainties. A uniform noise model was used, with only measurement noise (no process noise), and this noise was injected into the state space of the drone once we obtain it from the multirotor model in the simulation. The environment is assumed to be obstacle-free, allowing the study to focus solely on the effects of waypoint tracking, external disturbances, and system variations. The UAV is controlled through roll, pitch, and thrust, with yaw control omitted to reduce complexity while preserving stability. The rotational dynamics, including attitude control, are managed separately using the MATLAB UAV Toolbox multirotor model (version 2024a), which ensures stability by converting control inputs to PWM commands. This design choice is justified as AM applications primarily rely on precise lateral and vertical positioning rather than continuous yaw adjustments.  

To improve training efficiency and generalization, curriculum learning is introduced. The agent first learns basic waypoint navigation under controlled conditions before progressively handling more complex challenges, such as dynamic waypoints, mass variability, and external disturbances. This staged approach facilitates more stable learning and enhances adaptability. The UAV control problem is formulated as a MDP, The next section defines the MDP representation and details the integration of curriculum learning into the RL framework.

\section{Proposed Approach}

The proposed approach develops a DRL-based control system for UAV navigation in AM applications. The methodology is organized into two main phases:

Two off-policy DRL algorithms, DDPG and TD3, are evaluated for their performance in continuous action spaces. DDPG, while effective for high-dimensional tasks, is prone to overestimation bias, which TD3 mitigates using twin critics, target policy smoothing, and delayed updates. Both algorithms are tested in a simplified 3D simulation, where the UAV navigates between fixed waypoints. Metrics such as accuracy, precision, and success ratio are used for performance evaluation.

Next, curriculum learning is used to gradually increase task complexity. The agent starts with simple waypoint navigation in static environments and progresses to dynamic waypoints, variable payloads, and external disturbances. This staged approach stabilizes training, accelerates convergence, and ensures the UAV learns robust and adaptive control strategies for AM applications.

\subsection{System Architecture}

The environment is modeled as an MDP, characterized by state and action spaces. An MDP is defined by a tuple $(S,A,P,R,\gamma)$, where $S$ denotes the state space, $A$ represents the action space, $P: S \times A \times S \to [0,1]$ specifies the transition probability function $P(s'|s,a)$, indicating the likelihood of transitioning from the current state $s$ to a new state $s'$ upon executing action $a$; $\gamma$ is a discount factor within the range $(0,1)$, and $R: S \times A$ defines the reward function. The objective is to maximize cumulative rewards: 

\begin{equation}
\max_{\pi} \mathbb{E}\left[ \sum_{t=0}^{T} \gamma^t R_t \right],
\end{equation}

where \( \gamma \) is the discount factor, and \( R_t \) represents the reward at time \( t \), designed to ensure stability and precise trajectory tracking.

To enhance training efficiency and improve policy robustness, curriculum learning is implemented to gradually introduce complexity. The agent begins with basic navigation tasks and progressively encounters more challenging scenarios, including dynamic waypoints, variable mass distributions, and external disturbances. The task complexity at time step \( t \), represented as \( C(t) \), which evolves as \( C(t) \in \{C_1, C_2, \dots, C_n\} \). The curriculum-based objective is then defined as: 

\begin{equation}
\max_{\pi} \mathbb{E} \left[ \sum_{t=0}^{T} \gamma^t R(s_t, a_t, C(t)) \right],
\end{equation}

where \( C(t) \) adjusts task complexity throughout training, ensuring the UAV progressively learns to handle increasingly difficult conditions while maintaining stability. 

The state representation for the UAV in the proposed control framework is defined by a state vector $s_t$, which encapsulates key physical and positional parameters critical for navigation and control. This vector is expressed as: 

\begin{equation}
s_t = \begin{bmatrix}
a_{x,t}, & a_{y,t}, & a_{z,t}, & \Delta x_t, & \Delta y_t, & \Delta z_t, \\
v_{x,t}, & v_{y,t}, & v_{z,t}, & \theta_t, & \phi_t, & \psi_t, & z_t
\end{bmatrix},
\end{equation}

where $a_{x,t}, a_{y,t}, a_{z,t}$ are the drone's acceleration along the $x$-, $y$-, and $z$-axes, respectively. $\Delta x_t, \Delta y_t, \Delta z_t$ are the positional differences between the drone's current position and the target position along the $x$-, $y$-, and $z$-axes. $v_{x,t}, v_{y,t}, v_{z,t}$ are the velocity components of the drone in the $x$-, $y$-, and $z$-axes. $\theta_t, \phi_t$, and $\psi_t$ are the roll, pitch, and yaw angles, representing the drone's orientation around the $x$-, $y$-, and $z$-axes. $z_t$ is the current height of the drone.

To ensure consistent scaling and enhance the stability of the RL agent’s policy updates, each observation variable in $s_t$ is normalized using Z-score normalization:
\begin{equation}
s_t = \frac{s_t - \mu}{\sigma},
\end{equation}

where $\mu$ is the mean of each state variable and $\sigma$ is the standard deviation of each state variable.

The RL agent controls the multirotor using three actions: roll ($\phi_a$), pitch ($\theta_a$), and thrust ($T$). Yaw control is omitted in this configuration. Excluding yaw control reduces the complexity of the problem and allows the RL agent to stabilize and control the roll and pitch axes effectively. These actions are normalized to ensure compatibility with neural network policies and facilitate robust learning. The rotational dynamics, including attitude control, are managed by a separate algorithm using the MATLAB UAV Toolbox multirotor model (version 2024a), which handles the attitude rate control and converts the control inputs to PWM commands for stability.

\begin{figure}
    \centering
    \includegraphics[width=1.0\linewidth]{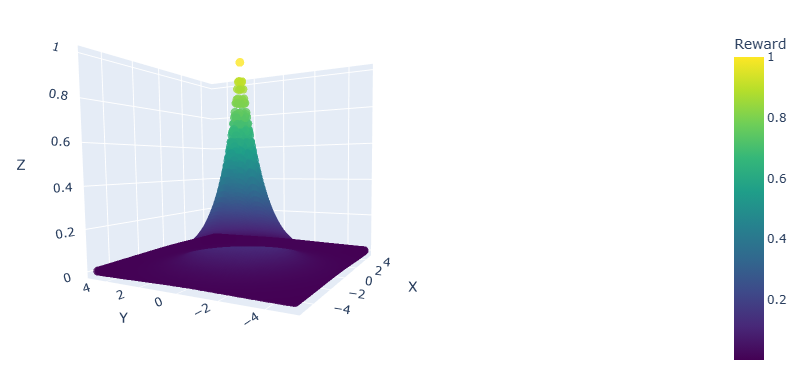} 
    \caption{3D Scatter Plot of Reward Function}
    \label{fig:reward-func}
\end{figure}

The reward function $r_p$ is designed to promote stable, precise navigation by rewarding proximity to the target position and discouraging positional deviations. Inspired by established reward-shaping techniques for positional control in robotics \cite{c21}, this convex reward function encourages the drone to progressively minimize its distance from the target. The reward function is defined as:
\begin{equation}
r_p = w_p \cdot  \exp\left(-\|p - p_{\text{desired}}\|^2\right),
\end{equation}

where $p$ represents the current position vector of the drone, $p_{\text{desired}}$ denotes the target position vector, $\|p - p_{\text{desired}}\|^2$ is the $L_2$ norm (Euclidean distance) between the current and target positions, and $w_p$ is a scaling factor to adjust the reward magnitude. By applying a negative exponential to the Euclidean distance, the function generates a reward that decreases smoothly as the drone moves away from the target. Experimental results and prior literature indicate that exponential reward functions can yield smoother, more stable trajectories in drone navigation tasks \cite{c21}, as shown in Fig. \ref{fig:reward-func}, which visualizes the convex reward distribution. In Fig. \ref{fig:reward-func}, the visualization of the reward function is presented for positional deviations of the drone along the X and Y axes, while keeping the Z-axis (height) constant. In this figure, the X and Y axes represent the drone's position in the horizontal plane, and the Z-axis illustrates the corresponding reward value that decays as the distance increases.

The transition function \( P(s_{t+1} \mid s_t, a_t) \) delineates how the drone transitions from its current state \( s_t \) to the next state \( s_{t+1} \) upon executing action \( a_t \) within the RL framework. In this study, the transition dynamics are modeled deterministically using MATLAB Simulink’s UAV Toolbox (version 2024a), which simulates the multi-rotor's physical behavior based on applied control inputs. The high-level actions are first scaled to match the UAV’s operational ranges and then translated into low-level Pulse-Width Modulation (PWM) \cite{c22} signals via a state estimator. The PWM commands directly influence the drone’s actuators as governed by the UAV’s dynamic model within Simulink. Consequently, the next state \( s_{t+1} \) is deterministically determined by the interplay between the current state, the scaled actions, and the inherent physical laws modeled in the simulation environment. 

The episode terminates when the drone reaches the target position with a velocity less than $0.1 m/s$ (near-zero velocity). Conversely, termination occurs under failure scenarios if the drone crashes by touching the ground (defined as a height less than 0.1 meters) or if it deviates more than 10 meters from the target position.

Two state-of-the-art off-policy DRL algorithms DDPG and TD3 are evaluated for their effectiveness in handling UAV navigation. Each algorithm is tested in simplified environments to determine baseline performance in waypoint navigation. For more information on the DDPG and TD3 algorithms, see \cite{c23, c24}.

\subsection{Curriculum Learning for Complexity Management}

To improve the robustness and adaptability of DRL agents, curriculum learning is employed to progressively increase task complexity. The task difficulty at time step $t$, represented as $C(t)$, transitions through predefined levels $C_1, C_2, \dots, C_n$, allowing the agent to build foundational skills before addressing more advanced challenges.

\textbf{Stages of curriculum learning include:}

\begin{itemize}
    \item \textbf{Basic Navigation ($C_1$):} The agent trains with static waypoints in a controlled environment, optimizing a reward function:
    \begin{equation}
    R(s, a) = -\|p_{\text{target}} - p_{\text{current}}\|_2,
    \end{equation}
    where \( p_{\text{target}} \) and \( p_{\text{current}} \) denote the target and current positions, respectively. 
    \item \textbf{Dynamic Waypoints ($C_2$):} Randomized start and target positions simulate adaptive path planning, with an added reward for waypoint completion.

    \item \textbf{Mass Variability ($C_3$):} Variable drone mass emulates material deposition, testing the agent's adaptation to center-of-gravity shifts.

    \item \textbf{External Disturbances ($C_4$):} Wind and noise test resilience, with penalties for deviations due to disturbances.
\end{itemize}

\section{Experiments}

\begin{figure}
    \centering
    \includegraphics[width=1.0\linewidth]{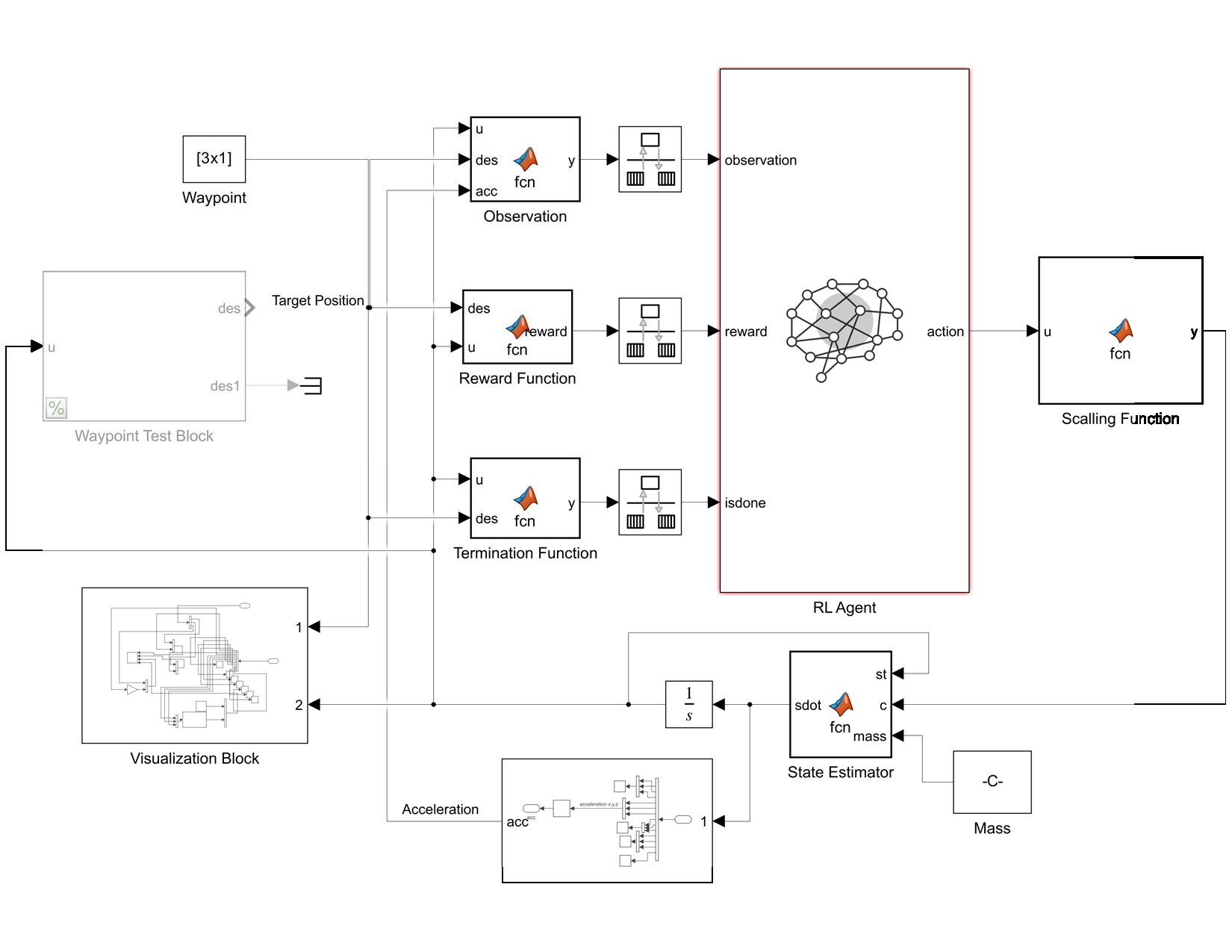}
    \caption{Simulink Model}
    \label{fig:simulink-model}
\end{figure}

This paper employs MATLAB’s Simulink platform (version 2024a), along with the RL Toolbox and UAV Toolbox, to develop a simulation environment for a drone engaged in AM. The Simulink setup shown in Fig \ref{fig:simulink-model} includes essential blocks such as the RL agent, observation, reward, and termination function, designed to enable real-time adaptive learning for stable control.

The Simulink environment functions in a continuous loop. At each step, the RL agent processes the latest observations to generate control actions. These actions are scaled, with roll and pitch constrained between \(-\frac{\pi}{2}\) and \(\frac{\pi}{2}\), and thrust limited between \(0\) and \(10\). The scaled actions are then processed through the state estimator and translated into commands for the UAV model. The drone’s updated state is integrated to produce the next observation, guiding the agent’s decisions in the subsequent step.

During the simulation, a dataset comprising state observations, actions, rewards, the subsequent state, and an episode termination flag is collected. This dataset forms the foundation for the RL agent’s learning process, enabling effective training through experience replay. Training is conducted using experience replay \cite{c25}, where the agent samples mini-batches from this dataset to update its policy and value function parameters. This approach mitigates the effects of temporal correlation between states, facilitating more stable learning. Experience replay also enhances training efficiency, allowing the agent to leverage past interactions to develop a robust policy capable of adaptive control in non-linear environments.

Table \ref{tab:hyperparameters} summarizes the hyperparameter settings for the DDPG and TD3 agents.

\begin{table}[t]
\caption{HYPERPARAMETER SETTINGS FOR DDPG AND TD3}
\centering
\renewcommand{\arraystretch}{1.4} 
\begin{tabular}{|p{4.0cm}|p{1.7cm}|p{1.7cm}|}
\hline
\textbf{Hyperparameter}               & \textbf{DDPG} & \textbf{TD3} \\
\hline
Optimizer                              & Adam          & Adam          \\
Learning Rate                          & $1 \times 10^{-3}$ & $1 \times 10^{-3}$ \\
Gradient Threshold                    & 1             & 1             \\
Sample Time                            & 0.01          & 0.01          \\
Discount Factor                        & 0.99          & 0.99          \\
Mini Batch Size                        & 256           & 256           \\
Experience Buffer Length               & $1 \times 10^{6}$ & $1 \times 10^{6}$ \\
Target Smooth Factor                   & $5 \times 10^{-3}$ & $5 \times 10^{-3}$ \\
Number of Epochs                       & 3             & 3             \\
Max Mini Batch per Epoch               & 100           & 100           \\
Learning Frequency                     & $-1$ (batch update) & $-1$ (batch update) \\
Policy Update Frequency                & 1             & 1             \\
Target Update Frequency                & 1             & 1             \\
Exploration Model                      & Gaussian Noise & Ornstein-Uhlenbeck Noise \\
\hline
\multicolumn{3}{|l|}{\textbf{Noise Parameters:}} \\
\hline
Mean Attraction Constant               & 1             & 1             \\
Noise Standard Deviation               & 0.1           & 0.1           \\
\hline
\multicolumn{3}{|l|}{\textbf{Target Policy Smooth Model:}} \\
\hline
Std. Dev. Min                            & -             & 0.05          \\
Std. Dev.                                & -             & 0.05          \\
Model Limits                           & -             & [$-0.5$, 0.5] \\
\hline
\end{tabular}
\label{tab:hyperparameters}
\end{table}

The neural network architectures of TD3 and DDPG share many similarities but are structured to align with their unique algorithmic needs. Both algorithms use an actor-critic framework with two fully connected layers in the actor network, consisting of 400 and 300 neurons. The primary distinction lies in the critic network. TD3 employs dual-critic networks, each with 400 and 300 neurons, to mitigate overestimation bias, while DDPG uses a single-critic network with the same structure.

For activation functions, both TD3 and DDPG utilize ReLU in the hidden layers and Tanh for the actor network's output to ensure smooth and bounded continuous action control.

In addition, each critic network is initialized using a weight scaling strategy \cite{datta2020survey}, where weights are sampled from a uniform distribution within the range 
\begin{equation}
\left[ -\frac{2}{\sqrt{\text{fan\_in}}}, \frac{2}{\sqrt{\text{fan\_in}}} \right].
\end{equation}

Here, \(\text{fan\_in}\) refers to the number of input units to the layer. This initialization ensures stable variance of activations across layers, preventing issues such as gradient vanishing or explosion during training. Actor networks are initialized using the same scaling method to maintain consistent training dynamics and facilitate efficient gradient flow.

\section{Results}

We conducted a comparative analysis of two RL algorithms, TD3 and DDPG, by training each agent on a target-reaching task. The objective was to navigate a drone to a specified target position. Evaluation metrics included average cumulative reward, standard deviation of rewards, average positional error measured in meters, precision (standard deviation of positional error), and success ratio across 100 test trials.

In the test phase, the drone always started from a fixed, stationary position, and the waypoints it needed to navigate to changed with each trial. In contrast, during training, both the drone's starting position and the waypoints were initialized randomly to expose the agent to a more diverse set of conditions. This approach helped the agent to become more robust by training in a variety of environmental configurations, ensuring its generalization ability across different scenarios.

\subsection{Performance on Target-Reaching Task}

\begin{figure}
    \centering
    \includegraphics[width=1.0\linewidth]{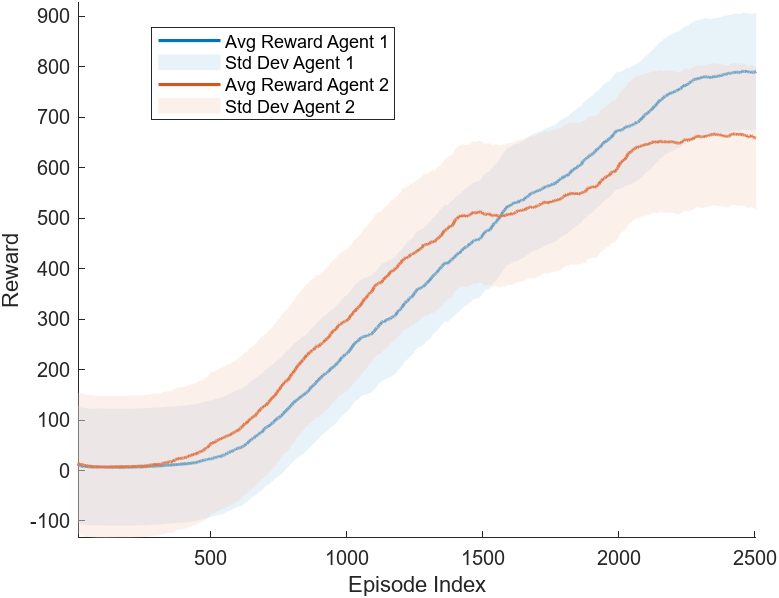}
    \caption{Training Comparison of TD3 and DDPG}
    \label{fig:test_td3_ddpg}
\end{figure}

The performance of TD3 and DDPG was analyzed using their training progress, as shown in Fig. \ref{fig:test_td3_ddpg}, where Agent 1 represents TD3 and Agent 2 represents DDPG. The TD3 agent outperformed DDPG, achieving a higher average cumulative reward, lower variance, and greater consistency in learning trends. A summary of the results over 100 tests is presented in Table~\ref{tab:target_reaching}.

\begin{table}[h!]
    \centering
    \renewcommand{\arraystretch}{1.4} 
    \caption{TEST RESULTS OF TD3 AND DDPG AGENTS}
    \label{tab:target_reaching}
    \begin{tabular}{|p{1.0cm}|p{1.0cm}|p{1.0cm}|p{1.cm}|p{1cm}|p{1.0cm}|}
        \hline
        \textbf{Agent} & \textbf{Avg. Cum. Reward} & \textbf{Std. Dev.} & \textbf{Avg. Pos. Error} & \textbf{Precision} & \textbf{Success (\%)} \\ \hline
        TD3 & 703 & 133.213 & $1.804 \times 10^{-2}$ & $2.26 \times 10^{-4}$ & 96 \\ \hline
        DDPG & 604 & 140.603 & $5.643 \times 10^{-2}$ & $6.412 \times 10^{-3}$ & 82 \\ \hline
    \end{tabular}%
\end{table}

TD3 achieved an average positional error of $0.01804m$, significantly lower than DDPG’s $0.05643m$, with higher precision, confirming its superior navigation performance. These results demonstrate TD3’s ability for precise and consistent target-reaching, making it the optimal agent for the subsequent waypoint navigation task.

\subsection{Waypoint Navigation Task}

To evaluate performance in more complex scenarios, the TD3 agent was tested on a waypoint navigation task involving multiple sequential targets. Without curriculum learning, TD3 achieved limited success, with lower cumulative rewards and success ratios. However, introducing curriculum learning, which progressively increases task complexity, significantly improved the agent’s performance.

As shown in Table \ref{tab:waypoint_navigation}, the curriculum-trained TD3 agent outperformed the non-curriculum-trained agent, achieving a higher average cumulative reward (\(729\) vs. \(593\)) and a greater success ratio (\(87\%\) vs. \(66\%\)). Additionally, the curriculum-trained agent exhibited a lower average positional error (\(0.01581m\)) compared to the non-curriculum-trained agent (\(0.01994m\)).

\begin{table}[h]
    \centering
    \renewcommand{\arraystretch}{1.4} 
    \caption{TRAINING RESULTS OF TD3 AGENT WITH AND WITHOUT CURRICULUM LEARNING}
    \label{tab:waypoint_navigation}
    \begin{tabular}{|p{2.8cm}|p{1.4cm}|p{1.2cm}|p{1cm}|}
        \hline
        \textbf{Agent} & \textbf{Avg. Cum. Reward} & \textbf{Std. Dev.} & \textbf{Success (\%)} \\ \hline
        TD3 (No Curriculum) & 593 & 145.9763 & 66 \\ \hline
        TD3 (With Curriculum) & 729 & 148.5167 & 87 \\ \hline
    \end{tabular}%
\end{table}

\begin{figure}
    \centering
    \includegraphics[width=1.0\linewidth]{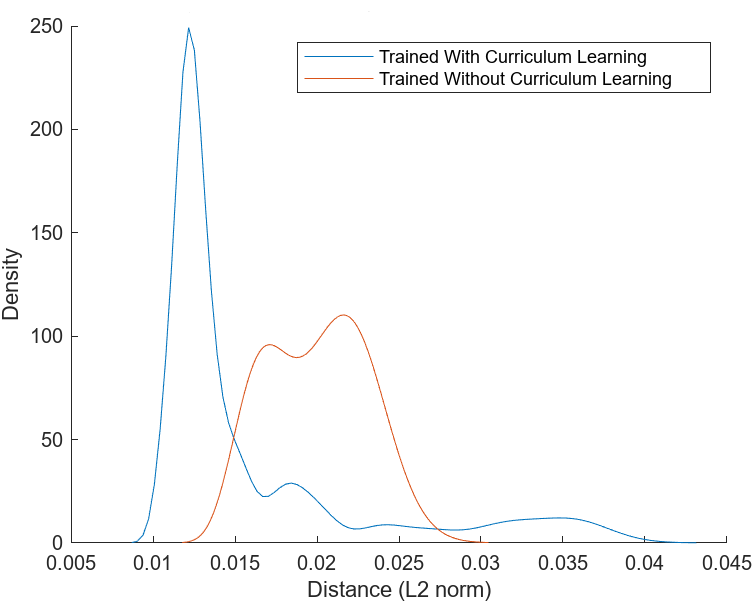}
    \caption{Test Results of TD3 Agent Trained with Different Methods}
    \label{fig:test-curr}
\end{figure}

The KDE (Kernel Density Estimate) plots of positional errors (Fig. \ref{fig:test-curr}) illustrate these improvements. The curriculum-trained agent’s error distribution is tightly clustered near zero, while the non-curriculum-trained agent exhibits a broader error spread, indicating greater variability. 

\subsection{Adaptation to Dynamic Mass Changes}

To address challenges arising from dynamic mass changes, such as those encountered during AM, the TD3 agent was adapted to account for variations in drone mass. Initially, the curriculum-trained agent achieved an average cumulative reward of $450$ and a success ratio of only $26\%$ under such conditions.

To enhance adaptability, the observation space was expanded to include the drone’s accelerations in the $x$, $y$ and $z$ directions. This modification allowed the agent to predict thrust commands corresponding to changes in mass, leveraging the relationship between thrust, mass and acceleration.

Retraining the agent with the expanded observation space led to significant improvements. The average cumulative reward increased from $454$ to $760$, and the success ratio rose from $29\%$ to $94\%$. Additionally, the average positional error was reduced from $0.2688 m$ to $0.09857 m$, and the precision (standard deviation of positional error), improved from $0.1459$ to $0.0626$.

\begin{figure}
    \centering
    \includegraphics[width=1.0\linewidth]{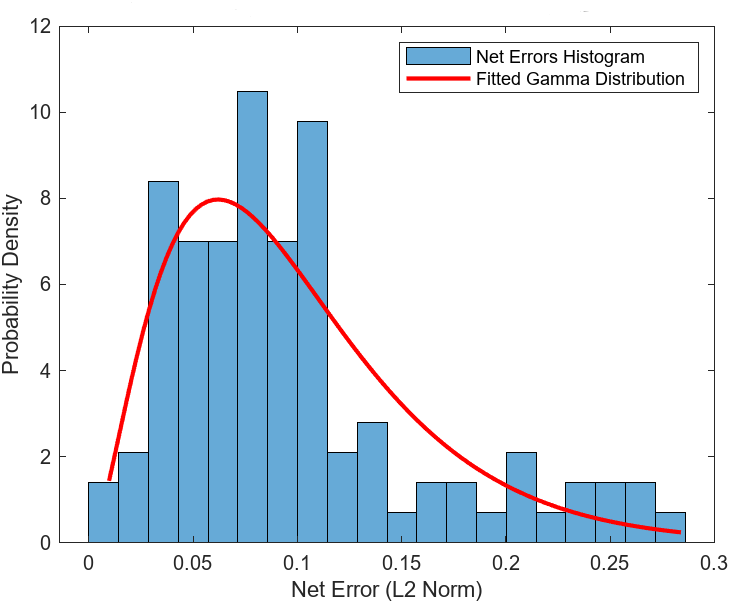}
    \caption{Test Results with Acceleration in Observation Space of TD3 Agent}
    \label{fig:test-dynamic-mass}
\end{figure}

\begin{figure}
    \centering
    \includegraphics[width=1.0\linewidth]{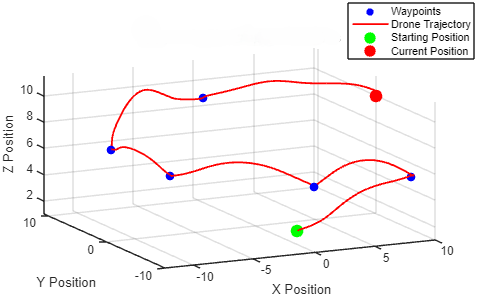}
    \caption{Final Demonstration of Drone Waypoint Navigation}
    \label{fig:trajectory}
\end{figure}

Fig. \ref{fig:test-dynamic-mass} presents a histogram of Euclidean distances between the drone’s final and target positions, fitted with a gamma distribution, while Fig. \ref{fig:trajectory} illustrates the path of the drone as it successfully navigates six randomaly generated waypoints under varying mass conditions.

\subsection{Summary of Agent Performance}

Table~\ref{tab:agent_summary} summarizes the training and testing results for all agents, highlighting their performance across various tasks. The TD3 agent demonstrated clear superiority in baseline tasks and in complex scenarios involving curriculum learning and dynamic mass adaptation. This progression underscores TD3’s robustness for real-world applications, such as waypoint navigation in environments with operational complexities.

 \begin{table*}[h!]
    \centering
    \renewcommand{\arraystretch}{1.4} 
    \caption{PERFORMANCE METRICS FOR DIFFERENT AGENTS ACROSS VARIOUS TRAINING ENVIRONMENTS}
    \label{tab:agent_summary}
    \begin{tabular}{|p{2.8cm}|p{2.2cm}|p{2.2cm}|p{1.4cm}|p{1cm}|p{1.4cm}|p{1.6cm}|p{1cm}|}
    \hline
    \textbf{Agent} & \textbf{Task Agent Trained On} & \textbf{Complexity of Environment} & \textbf{Avg. Cum. Reward} & \textbf{Std. Dev. of Reward} & \textbf{Avg. Pos. Error (in meter)} & \textbf{Precision (Std. Dev. of Pos. Error)} & \textbf{Success Ratio (\%)} \\ \hline
    DDPG & Single Waypoint Navigation & No Noise Added; No Variable Mass & 604 & 140.6029 & $5.643 \times 10^{-2}$ & $6.412 \times 10^{-3}$ & 82 \\ \hline
    TD3 & Single Waypoint Navigation & No Noise Added; No Variable Mass & 703 & 133.2130 & $1.804 \times 10^{-2}$ & $2.26 \times 10^{-4}$ & 96 \\ \hline\hline
    TD3 (No Curriculum Learning) & Multiple Waypoint Navigation & Noise Added; No Variable Mass & 593 & 145.9763 & $1.994 \times 10^{-2}$ & $2.780 \times 10^{-3}$ & 66 \\ \hline
    TD3 (With Curriculum Learning) & Multiple Waypoint Navigation & Noise Added; No Variable Mass & 729 & 148.5167 & $1.581 \times 10^{-2}$ & $6.831 \times 10^{-3}$ & 87 \\ \hline\hline
    TD3 (With Acceleration in Observation Space) & Multiple Waypoint Navigation & Noise Added; Variable Mass & 760 & 152.5167 & $9.857 \times 10^{-2}$ & $6.260 \times 10^{-2}$ & 94 \\ \hline
    \end{tabular}
\end{table*}


\section{Conclusion}

This study evaluated the performance of two RL algorithms, DDPG and TD3, for UAV control in additive manufacturing tasks. Results showed that TD3 outperformed DDPG in key metrics, including cumulative reward, average positional error, precision, and success ratio, confirming its suitability for precise and consistent target-reaching tasks. Given these results, TD3 was selected for further evaluation in more complex tasks, as it showed better performance compared to DDPG in the initial tests.

The integration of curriculum learning further enhanced TD3’s performance, enabling the agent to handle complex waypoint navigation scenarios. Curriculum-trained TD3 achieved higher rewards, greater success ratios, and improved accuracy compared to its non-curriculum counterpart. Additionally, adapting the observation space to include acceleration data allowed TD3 to handle dynamic mass variations effectively, achieving a cumulative reward of 760 and a success ratio of 94\%.

Overall, TD3 demonstrated superior adaptability and robustness across all tested scenarios, including dynamic environments with noise and mass changes. These findings highlight its potential for real-world AM applications, providing a robust framework for UAV waypoint navigation under operational complexities. Future work will focus on real-world implementation, and expanding the agent’s capabilities for diverse AM scenarios.

\end{document}